\def\year{2020}\relax
\documentclass[letterpaper]{article} 
\usepackage{aaai20}  
\usepackage{times}  
\usepackage{helvet} 
\usepackage{courier}  
\usepackage[hyphens]{url}  
\usepackage{graphicx} 
\usepackage{graphicx}  
\title{Adversarial Image Translation: \\
Unrestricted Adversarial Examples in Face Recognition Systems}
\author{Kazuya Kakizaki\textsuperscript{\rm 1} and Kosuke Yoshida\textsuperscript{\rm 1}\thanks{both authors contributed equally}\\ 
\textsuperscript{\rm 1}NEC Corporation\\ 
7-1, Shiba 5-chome Minato-ku, Tokyo 108-8001 Japan\\
kazuya1210@nec.com, dep58@nec.com}

\begin{document}
\maketitle
\begin{abstract}
Thanks to recent advances in deep neural networks (DNNs), face recognition systems have become highly accurate 
in classifying a large number of face images. 
However, recent studies have found that DNNs could be vulnerable to adversarial examples, raising concerns about the robustness 
of such systems. 
Adversarial examples that are not restricted to small perturbations could be more serious since conventional certified defenses 
might be ineffective against them. 
To shed light on the vulnerability to such adversarial examples, we propose a flexible and efficient method for generating 
unrestricted adversarial examples using image translation techniques. 
Our method enables us to translate a source image into any desired facial appearance with large perturbations to deceive target face 
recognition systems. Our experimental results indicate that our method achieved about $90$ and $80\%$ attack success rates 
under white- and black-box settings, respectively, and that the translated images are perceptually realistic and maintain 
the identifiability of the individual while the perturbations are large enough to bypass certified defenses.
\end{abstract}

\section{Introduction}
Deep neural networks (DNNs) have become more accurate than humans in image classification~\cite{krizhevsky2012imagenet} and 
machine translation~\cite{bahdanau2014neural}. 
However, recent studies have shown that DNNs could be vulnerable to adversarial examples
~\cite{Szegedy2014IntriguingPO,Ian2015expaining,carlini2017towards}. 
Specifically, DNNs could be intentionally deceived by an input data point that has been slightly modified. 
To understand the mechanism of such a vulnerability and make DNNs more robust, it is important to study methods for generating 
adversarial examples.

The vulnerability to adversarial examples raises concerns about face recognition systems widely used in applications such as 
biometric authentication and public safety~\cite{masi2018deep}. 
Since the recent success of face recognition systems rely on DNNs, a potential attacker could exploit adversarial examples 
for incorrect recognition or impersonating another person. 
Therefore, it is important to consider the risk of adversarial examples attacking face recognition systems.

\begin{figure}[t]
\begin{center}
    \begin{tabular}{c}                   
      \begin{minipage}{0.25\hsize}
        \begin{center}
          \includegraphics[width=0.8\columnwidth,natwidth=250,natheight=700]{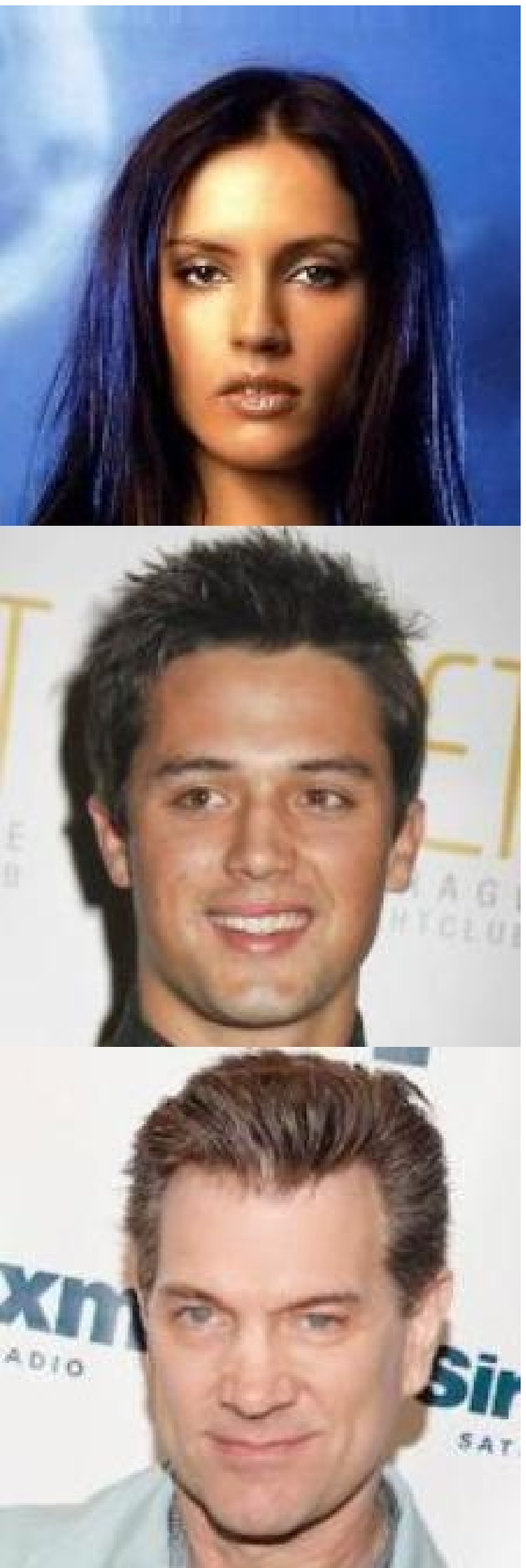}
          (a)Original 
        \end{center}
      \end{minipage}
      \begin{minipage}{0.25\hsize}
        \begin{center}
          \includegraphics[width=0.8\columnwidth,natwidth=250,natheight=700]{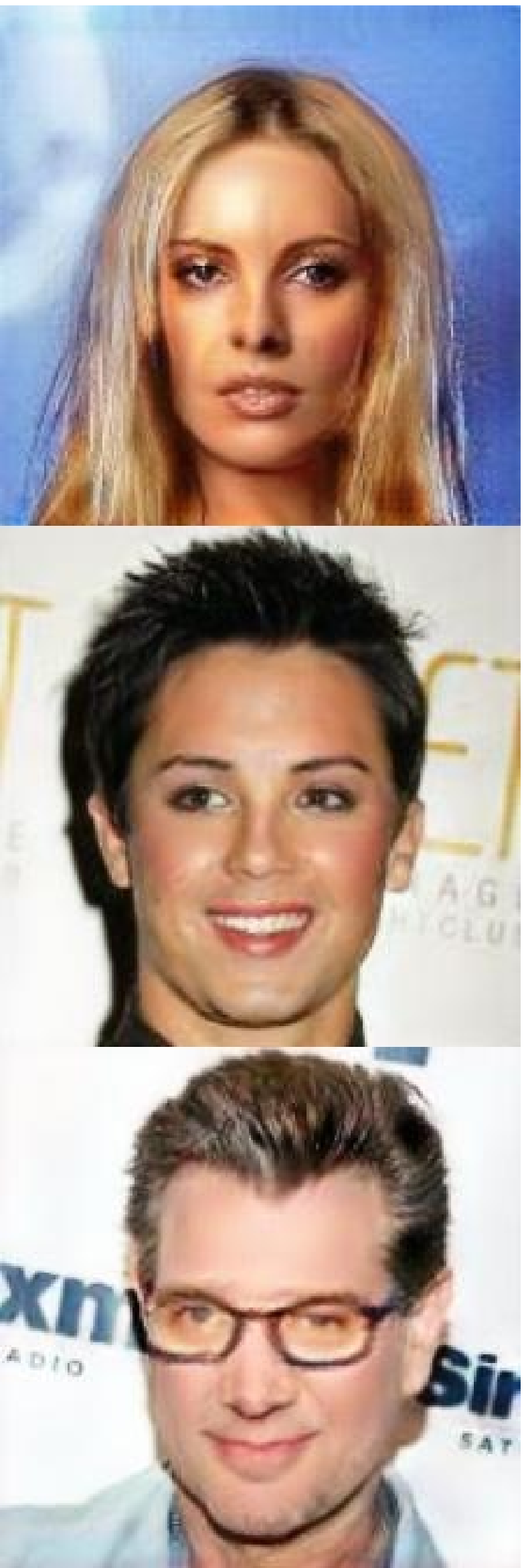} 
          (b)Adversarial 
        \end{center}
      \end{minipage}
      \begin{minipage}{0.25\hsize}
        \begin{center}
          \includegraphics[width=0.8\columnwidth,natwidth=250,natheight=700]{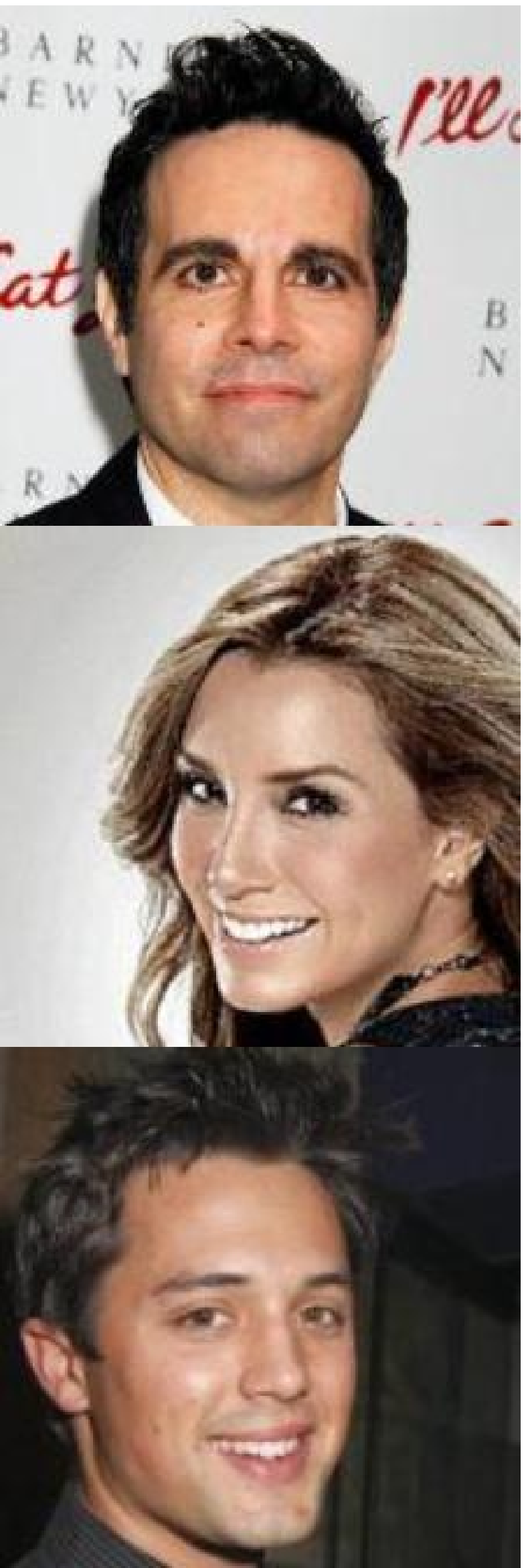}
          (c)Target
        \end{center}
      \end{minipage}      
    \end{tabular}
    \caption{Adversarial image translation against face recognition systems. 
    Our method translates (a) original images into (b) adversarial images with desired domain labels to classify them as (c) target images. 
    Corresponding domain labels are blond hair, makeup, and eyeglasses from top to bottom.}
    \label{fig:Intro}
\end{center}
\end{figure}

Typical methods for creating adversarial examples for deceiving target classifiers involve adding small carefully crafted 
perturbations on a source image~\cite{Szegedy2014IntriguingPO,Ian2015expaining,carlini2017towards}. 
To reduce the risk of adversarial examples based on small perturbations, several defenses have been proposed with theoretical 
certification. These defenses provide a lower bound of class-changing perturbations based on the global Lipschitz constant of 
DNNs~\cite{cisse2017parseval,gouk2018regularisation,NIPS2018_7889} and 
randomized smoothing~\cite{li2018second,cohen2019certified,lecuyer2018certified}. 
These studies indicate that small-perturbation-based attacks may no longer be effective with these defenses.

In spite of these certified defenses, there remains a significant risk of adversarial examples. 
Song et al. and Brown et al. independently introduced a novel concept called unrestricted adversarial examples
~\cite{NIPS2018_8052,brown2018unrestricted}. 
While such adversarial examples are not restricted to small perturbations, they do not confuse human observers. 
For instance, an image of a stop sign is still recognized as a stop sign by human observers, though the adversarial perturbations are 
large enough to bypass certified defenses. 
These adversarial examples could be a serious security issue and shed light on the mechanism of this type of vulnerability.

In the context of face recognition systems, how can we define unrestricted adversarial examples in a reasonable scenario? 
Spatial transformation, such as adversarial rotation~\cite{brown2018unrestricted} and distortion~\cite{goswami2018unravelling}, 
are impractical for attacks against face recognition systems since adversaries are unable to control the spatial transformations 
in biometric authentication and public surveillance (e.g. rotated images cannot be presented in a passport-control scenario). 
Therefore, we focus on unrestricted adversarial examples without spatial transformation. 
We consider the following conditions: our adversarial examples ($1$) should be perceptually realistic enough to maintain 
the identifiability of the individual in the original image and ($2$) have large enough perturbations 
to bypass defenses that are based on small perturbations.

To generate unrestricted adversarial examples that satisfy these two conditions, we propose a method that takes advantage of recent 
image translation techniques into different domains~\cite{choi2018stargan}. 
Our method enables us to translate the facial appearance in a source image into several domains so that face recognition systems can 
be deceived. 
While the translation provides perceptible perturbations on the face, it avoids damaging the identifiability of the individual 
of the source for human observers.

Figure~\ref{fig:Intro} illustrates the flow of our method: (a) the original images are translated into (b) adversarial images 
with respect to several domains to be classified as (c) the target images. From top to bottom, we translate in three different domain 
labels: blond hair, makeup, and eyeglasses. 
They show that changing hair-color and adding face accessories, such as makeup and eyeglasses, deceive face recognition systems. 
Translating a given image into any desired facial appearance in this way enables us to evaluate the risk of diverse unrestricted 
adversarial examples.

In our experiments, we found that our method achieved about $90\%$ attack success rate on average against publically available face 
recognition models in VGGFace and its latest version VGGFace2. 
Our method can also generate realistic adversarial images with the desired facial appearance and maintain the identifiability of the individual 
even with large perturbations. 
We applied our method to a black-box setting and evaluated black-box attacks based on the transferability phenomenon and dynamic 
distillation strategy~\cite{Szegedy2014IntriguingPO,ijcai2018-543}.

Our contributions are listed as follows.

\begin{itemize}
 \item We propose a flexible and efficient method for generating unrestricted adversarial examples against face recognition systems.
\item We confirm that the generated adversarial examples are perceptually realistic enough to maintain the identifiability of the individual to avoid confusing human observers.
\item We experimentally demonstrated that our method can deceive face recognition systems with high attack success rates under white- and black-box settings. 
\item We confirm that generated adversarial examples bypass a state-of-the-art certified defense.

\end{itemize}

This paper is organized as follows. 
In Section 2, we review related studies on image translation methods based on generative 
adversarial networks (GANs) and adversarial examples in face recognition systems. 
In Section 3, we provide details of our proposed method. 
In Section 4, we report the settings and results of our experiments. 
We conclude this paper in Section 5.

\section{Related Studies}
In this section, we review several related studies on unrestricted adversarial examples, GANs, and adversarial face accessories.

\subsection{Unrestricted Adversarial Examples}
Song et al. proposed a method for generating unrestricted adversarial examples from scratch instead of adding small perturbations 
on a source image and demonstrated that their generated examples successfully bypassed several certified defenses that are based on 
small perturbations~\cite{NIPS2018_8052}. 
They adopted two classifiers: a target classifier that they wish to deceive and an auxiliary classifier that provides correct predictions. 
These classifiers encourage the examples generated from scratch to deceive the target classifier without changing any semantics. 
They confirmed that their method successfully created unrestricted adversarial examples without confusing human observers 
by using Amazon Mechanical Turk (AMTurk). 
While their work is notable, more effort is required to deceive face recognition systems.

\subsection{Generative Adversarial Networks (GANs)} 
GANs have achieved significant results especially in image generation tasks
~\cite{goodfellow2014generative,karras2018progressive,karras2018style}. 
They consist of two components: a generator and discriminator. 
The generator is trained to provide fake images that are indistinguishable from real ones by the discriminator, 
while the discriminator is trained to distinguish fake images from real images. 
This competitive setting is represented by an adversarial loss.

Xiao et al. proposed a method for exploiting GANs to generate realistic adversarial images~\cite{ijcai2018-543}. 
With their method, the generator is trained to provide adversarial images that are indistinguishable from real ones by the 
discriminator. They demonstrated that their generated adversarial examples were perceptually realistic through human evaluation. 
The generator also efficiently provides adversarial examples once it is trained. 
This could be beneficial to potentially improve the robustness of target models.

Choi et al. introduced a framework called StarGAN that enables us to translate an input image into multiple domains using GANs~\cite{choi2018stargan}. 
In addition to conventional GANs, they introduced two loss functions: auxiliary classification loss and reconstruction loss. 
The first one is to guarantee that the output image can be classified into the corresponding domain label~\cite{odena2017conditional}. The second one is to preserve the content of the input image in the translated image as cycle consistency loss~\cite{zhu2017unpaired} since StarGAN formulation consists of only a single generator. They demonstrated that StarGAN is useful in translating face images into any desired domain or facial expression in a flexible manner.

\subsection{Deceiving Face Recognition Systems} 
Some studies modified the facial appearances of source images by adding adversarial face accessories and deceived face recognition 
systems for the purpose of impersonation or privacy preservation~\cite{sharif2016accessorize,sharif2017adversarial,feng2013facilitating}. 
For instance, Sharif et al. created adversarial eyeglasses by iteratively updating their color. 
These eyeglasses allow adversaries to impersonate another person with high success rates in several state-of-the-art face 
recognition systems. 
While these studies demonstrated the significant risk of adversarial examples, they have limited scalability to understand the mechanism of such vulnerability.

We take advantage of recent studies on image translation with GANs to translate the facial appearance of source images 
in an adversarial manner. 
This translation introduces large perturbations on the source images; therefore, translated images are unrestricted adversarial 
examples against face recognition systems.

\begin{table*}[t]
\caption{Accuracy of target face recognition models on legitimate datasets, evaluated using images held aside for testing. 
Our target face recognition models achieved high accuracy in all cases. VGG(A) and ResNet(A): VGG16 and ResNet50 for case A. 
VGG(B) and ResNet(B): VGG16 and ResNet50 for case B. 
StarGAN: image translation without adversarial effect ($\lambda_{\gamma}=0$).}
\begin{center}
\begin{tabular}{l|cccc}
           &  VGG(A) & ResNet(A) & VGG(B) & ResNet(B)        \\ \hline
  legitimate &  0.97   & 1.00       & 0.95   & 0.96          \\
  StarGAN~\cite{choi2018stargan} & 0.85 & 0.93 & 0.73 & 0.75 \\	
\end{tabular}
\label{tab:legi}
\end{center}
\end{table*}
\section{Our Method}
We first provide problem definitions and notations for our method then introduce our formulation to translate hair color, makeup, 
and eyeglasses of a people’s facial images so that the target face recognition models can be deceived from the impersonation of 
others.

\subsection{Problem Definition}
Given an instance $(x_{i}, y_{i}, c_{i})$, which is composed of a face image $x_{i} \in \cal{X}$ sampled according 
to some unknown distribution, class label $y_{i} \in \cal{Y}$ corresponding to the person of the image, and domain label $c_{i}$ representing the 
existence of each binary domain. 
Here, $c_{i}$ is a binary vector whose $j$-th component is $1$ when the corresponding image exhibits the $j$-th binary domain. 
The target face recognition models learn a classifier $\phi:\cal{X} \rightarrow \cal{Y}$ that assigns class labels into each face image. Our objective is to generate adversarial example $\hat{x_{i}}$ classified as $\phi(\hat{x_{i}}) = t\: (t \neq y_{i})$, where $t$ is our target class label. 
In addition, $\hat{x_{i}}$ should have the desired domain label.

\subsection{Formulation}
Our method is mainly based on a framework of recent GANs and consists of four components: a generator $G$, discriminator $D_{s}$, auxiliary classifier $D_{c}$, and target model $\phi$. The $G$ provides images indistinguishable for the $D$ by optimizing an adversarial loss $L_{wgan}$. The generated images are encouraged to have any desired domain label by minimizing a classification loss $L^{r}_{cla}, L^{f}_{cla}$. The target loss $L_{tar}$ is minimized to deceive the target models (i.e. face recognition systems) through impersonation.

Adopted from StarGAN~\cite{choi2018stargan}, we train a single $G$ to translate input face images into output images with any 
desired domain label. 
The $G$ takes a face image $x$ and a desired domain label $c_{out}$ as an input and generates an output image $\hat{x}$ 
whose domain label is $c_{out}$, $G:x, c_{out} \rightarrow \hat{x}$. 
The $D_{s}$ takes $\hat{x}$ and provides a probability distribution over sources, $D_{s}(\hat{x})$. 
The goal with these components is to optimize the adversarial loss defined as 
\begin{equation}
L_{gan} = E_{x}[\log D_{s}(x)] + E_{x}[\log(1-D_{s}(G(x,c_{out})))],
\end{equation}
where the $G$ attempts to minimize the loss while the $D_{s}$ attempts to maximize it to generate realistic images.

To take advantage of recent techniques for stabilizing GAN training, we adopt the Wasserstein GAN with gradient 
penalty~\cite{gulrajani2017improved} defined as
\begin{eqnarray}
L_{wgan} &=& E_{x}[D_{s}(x)] + E_{x}[(1-D_{s}(G(x,c_{out})))] \nonumber \\
&&\qquad\qquad  -\lambda_{wgan}L_{pen},
\end{eqnarray}
where $\lambda_{wgan}$ represents a hyper parameter that controls the magnitude of the penalty term $L_{pen}$. 
To enforce the Lipschitz constant, the penalty term is defined as
\begin{equation}
L_{pen} = E_{x}[(\Vert \nabla_{x_{q}} D_{s}(x_{q}) \Vert_{2} - 1)^{2}],
\end{equation}
where $x_{q}$ is sampled uniformly along straight lines between real and generated images. We set $\lambda_{wgan}$ to 10.

To encourage the generated image to have the desired domain label $c_{out}$, 
we adopt the auxiliary classification loss~\cite{odena2017conditional}. 
Specifically, we optimize the $D_{c}$ to classify the real images as corresponding domain labels by minimizing the loss, which is defined as
\begin{equation}
L_{cla}^{r} = E_{x, c_{in}}[-\log D_{c}(c_{in} | x )],
\end{equation}
where $D_{c}(c_{in} | x)$ represents a probability distribution over the domain labels and $c_{in}$ corresponds to the domain label of an input. 
We then optimize the $G$ to classify the generated images as any target domain label by minimizing the loss, which is defined as
\begin{equation}
L_{cla}^{f} = E_{x, c_{out}}[-\log D_{c}(c_{out} | G(x,c_{out}))].
\end{equation}

Adopted from a common practice of image translation to maintain the identifiability of the individual~\cite{zhu2017unpaired}, 
we add reconstruction loss, which is defined as 
\begin{equation}
L_{rec} = E_{x, c_{out}, c_{in}}[\Vert x - G(G(x,c_{out}), c_{in}) \Vert_{1}].
\end{equation}
Note we adopt L1 norm to obtain less blurring images.

We add loss to encourage a target model $\phi$ to classify the generated images as target labels $t$. 
This loss is defined as
\begin{equation}
L_{tar} = E_{x} [\max (\max_{i \neq t} z_{i}(x) - z_{t}(x), \kappa)],
\end{equation}
where $z$ is the output of $\phi$ except the final softmax layer (i.e. logits) and $\kappa$ is a hyper parameter set to negative values in our experiments.

Finally, our full loss function is defined as
\begin{eqnarray}
L_{G} &=& L_{wgan} + \lambda_{\alpha} L_{cla}^{f} + \lambda_{\beta} L_{rec} + \lambda_{\gamma} L_{tar},\\
L_{D} &=& - L_{wgan} + \lambda_{\alpha} L_{cla}^{r}, 
\label{eq:full}
\end{eqnarray}
where we obtain our $G$, $D_{s}$, and $D_{c}$ by minimizing $L_{G}$ and $L_{D}$, respectively. 
Note $\lambda_{\alpha}$, $\lambda_{\beta}$, and $\lambda_{\gamma}$ are hyper parameters that control the relative importance of 
each loss function.

\begin{table*}[tbph]
\caption{Average, maximum, and minimum attack success rates. Our generated adversarial examples achieved about $90\%$ attack success 
rate on average against two models with different domain labels.}
\begin{center}
\scalebox{1.0}[1.0]{
\begin{tabular}{l|cccc|cccc|cccc}
          &\multicolumn{4}{|c|}{hair color (black/blond)} &\multicolumn{4}{|c|}{makeup} &\multicolumn{4}{|c}{eyeglasses}\\  \hline
          &\multicolumn{2}{|c|}{VGG}&\multicolumn{2}{|c|}{ResNet}&\multicolumn{2}{|c|}{VGG}&\multicolumn{2}{|c|}{ResNet}&\multicolumn{2}{|c|}{VGG}&\multicolumn{2}{|c}{ResNet}\\ \hline
          &A   &B   &A   &B   &A   &B   &A   &B   &A   &B   &A   &B    \\ \hline
    Ave.  &0.85&0.98&0.95&0.96&0.83&0.98&0.94&0.96&0.83&0.96&0.93&0.86 \\ 
    Max.  &0.95&0.99&1.00&0.98&0.90&0.99&1.00&0.98&0.94&0.98&1.00&0.97 \\ 
    Min.  &0.70&0.97&0.75&0.92&0.65&0.96&0.66&0.94&0.74&0.94&0.78&0.50 \\ 	
\end{tabular}
}
\end{center}
\label{tab:asrA}
\end{table*}

\section{Experiments Results}
We first demonstrated that our method achieves more than a $90\%$ attack success rate 
against two target models in a white-box setting with high quality images. 
We then evaluated black-box attacks based on the transferability phenomenon of adversarial examples~\cite{Szegedy2014IntriguingPO,Ian2015expaining,papernot2016practical} and 
the dynamic distillation strategy~\cite{ijcai2018-543}. 
Finally, we show that adversarial examples obtained with our method satisfy two conditions: they are (1) realistic enough to maintain 
the identifiability of the individual and (2) have large perturbations to bypass certified defenses.

\subsection*{Dataset}
We used the CelebA dataset, which has 202,599 face images with 40 binary attributes~\cite{liu2015faceattributes}. 
To train the target face recognition models, we randomly selected $10$ persons as case A and $100$ persons as case B. 
We only used $20$ images per person for the training dataset and the rest of the images were held aside for the test dataset.

\subsection*{Target Face Recognition Models}
Our target face recognition models were VGGFace~\cite{Parkhi15} and VGGFace2~\cite{Cao18}.
We downloaded publically available pre-trained VGG16~\footnote[1]{https://github.com/yzhang559/vgg-face} for VGGFace and 
ResNet50~\footnote[2]{https://github.com/rcmalli/keras-vggface} for VGGFace2. 
These pre-trained models exhibit state-of-the-art results for face recognition tasks. 
They take a 224x224 face image as an input and provide a low-dimensional face descriptor in which two images of the same person are designed to be closer to each other.

On top of the pre-trained face descriptor, we constructed a fully connected layer to define our own target face recognition models 
with the CelebA dataset. 
We fine-tuned the parameters from the fully connected layer using categorical cross entropy loss with the training dataset. 
Our fine-tuned models achieved more than $95\%$ accuracy on the test dataset, as shown in Table~1. 
This table also shows the test accuracy on translated images without any adversarial effect: in which $\lambda_{\gamma} = 0$ in 
Equation~(8), indicating that the models are robust against simply changing domain labels in StarGAN.

\subsection*{Implementation Details}
We adopted a similar architecture from image translation studies~\cite{zhu2017unpaired,choi2018stargan}. 
In particular, we constructed two down-sampling blocks, six residual blocks, and two up-sampling blocks with rectified linear unit 
(ReLU) layers for the $G$, and five down-sampling blocks with leaky ReLUs and fully connected blocks for the $D$. 
We applied PatchGAN~\cite{isola2017image} to construct the $D$ following previous studies.

We trained both models using Adam with $\beta_{1} = 0.5$ and $\beta_{2} = 0.999$. 
The learning rate linearly decreased~\cite{choi2018stargan} and batch size was set to $32$ in all experiments. 
We used $\lambda_{\alpha} = 1$, $\lambda_{\beta} = 10$ for $L_{G}$ and $L_{D}$ and $\kappa = -0.3$ for $L_{tar}$. 
For cases A and B, we set $\lambda_{\gamma}$ to $0.2$ and $0.5$ and number of epochs to $300,000$ and $500,000$, respectively.

\subsection{Attacks in White-Box Setting}
We first evaluated our method under a white-box setting in which the adversary has access to the model architecture and its weights. 
Since we chose all individuals as targets of impersonation, we evaluated ten different $G$s and $D$s for case A. 
For case B, we randomly chose 5 of the 100 individuals as targets of impersonation. 
Our evaluation of the attack success rate involved using the test dataset held aside from training of our models for fair evaluation.

Theoretically, we can apply 40 types of domain labels available in the CelebA dataset. 
However, we only applied three domain labels: hair color (black/blond), heavy makeup, and eyeglasses since our goal was to efficiently
 generate images in practical scenarios for attacking face recognition systems~\cite{sharif2016accessorize,sharif2017adversarial,feng2013facilitating}.

We report the average, maximum, and minimum attack success rates among different models in both cases. 
Table~\ref{tab:asrA} shows that our method achieved high attack success rates in both models and domain labels. 
From the performance results in Table~\ref{tab:asrA}, we can see that about $90\%$ of test images could be successful in
deceiving our target face recognition models by changing their facial appearance.

Figure~3 illustrates successful adversarial examples. 
The left columns show the original images and middle ones show the adversarial examples with several domain labels: 
blond/black hair, makeup, and eyeglass from top to bottom. 
The target images that we tried to impersonate are presented in the right columns. 
These images demonstrate that our method accurately translates source images into multiple domains, enabling 
the target face recognition models to be deceived. 
Moreover, the introduced perturbations are perceptible to human observers while avoiding the damages in the identifiability of the individual 
of the source images.

\begin{table*}[t]
\caption{Average attack success rates in transferability-based attacks between two target face recognition models. 
About $30\%$ of attacks transferred from VGG16 to ResNet50, and vice versa.}
  \begin{center}
  \begin{tabular}{l|l|cc} 
Model         & Domain              & VGG(A/B)    & ResNet(A/B)  \\ \hline
              & hair color          & -           & 0.26/0.28    \\ \cline{2-4}
VGG(A/B)      & makeup      	    & -           & 0.22/0.33    \\ \cline{2-4}
              & eyeglasses          & -           & 0.27/0.36    \\ \hline 
              & hair color          & 0.35/0.32   & -            \\ \cline{2-4}
ResNet(A/B)   & makeup              & 0.29/0.30   & -            \\ \cline{2-4}
              & eyeglasses          & 0.22/0.38   & -            \\ 
  \end{tabular}
  \end{center}
\end{table*}

\begin{figure*}[t]
    \begin{tabular}{lr}
                   
      \begin{minipage}{0.5\hsize}
        \begin{center}
          \includegraphics[width=0.5\columnwidth,natwidth=300,natheight=314]{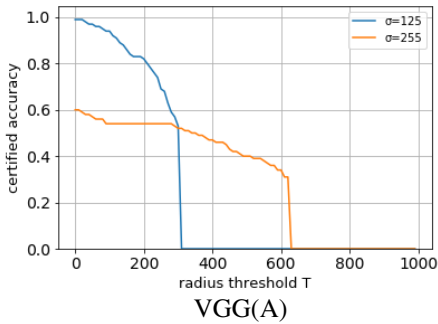}
          
          \hspace{2cm}VGG(A)
        \end{center}
      \end{minipage}
      \begin{minipage}{0.5\hsize}
        \begin{center}
          \includegraphics[width=0.5\columnwidth,natwidth=300,natheight=314]{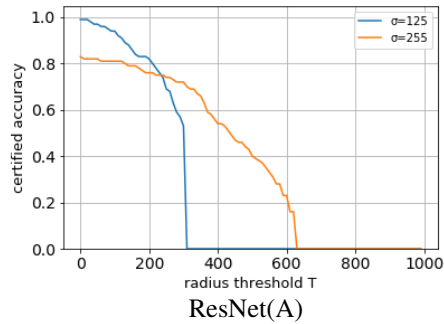} 
          
          \hspace{2cm}ResNet(A)
        \end{center}
      \end{minipage}
      
    \end{tabular}
    \caption{Certified accuracy and radius threshold $T$ for VGG(A) (left) and ResNet(A) (right). 
    Horizontal axis represents $T$ and vertical axis represents certified accuracy: proportion of images classified correctly and 
    whose certified radius is less than $T$. 
    In both settings, $\sigma = 125$ and $255$ respectively, and certified radii were much less than perturbations introduced with 
    our method.}
    \label{fig:random_smooth}
\end{figure*}

\subsection{Attacks in Black-Box Setting}
Since most face recognition systems do not allow anyone to acquire knowledge about their network architectures and weights, 
it is important to analyze vulnerabilities in black-box setting. 
We explored the attack strategies based on the transferability phenomenon in which adversarial examples generated in a model 
will also lead to successful attacks against other models~\cite{Szegedy2014IntriguingPO,Ian2015expaining,papernot2016practical}. 
We generated adversarial images with one target face recognition model and evaluated the attack success rates against another model.

Table~3 lists the results of transferability-based attacks between the two target face recognition models: VGG16 and ResNet50. 
We found that about $30\%$ of attacks transferred from VGG16 to ResNet50, and vice versa. 
These results indicate that publically available face recognition systems are vulnerable to simple transferability-based attacks 
even without any knowledge about the models.

We also evaluated our method based on the dynamic distillation strategy in a black-box setting~\cite{ijcai2018-543}. 
This enables us to obtain a distilled model that behaves similar to a black-box model. 
As described in previous studies, we repeated the following two steps in each iteration.

\textbf{First step}. Update the distilled model $\hat{\phi}$ with a fixed $G$ and $D$ by minimizing the following objectives:
\begin{equation}
E_{x} [H(\hat{\phi}(x),b(x))] + E_{x}[H(\hat{\phi}(G(x, c_{out})), b(G(x, c_{out}))]
\end{equation}
where $H$ denotes the cross-entropy loss and $\hat{\phi}(x)$ and $b(x)$ denote the output of the distilled model and black-box model, respectively. This step encourages the distilled model to behave similar to the target black-box model on generated adversarial examples.

\textbf{Second step}. Update our $G$ and $D$ with $\hat{\phi}$ using Equations (8) and (9). In this step, we can uses our method on the distilled models, as described in a white-box setting.

Table~4 shows that with the dynamic distillation, our method achieved a more than $80\%$ attack success rate for VGG16 and about 
$70\%$ for ResNet50 on average with $200,000$ epochs. 
These results indicate that the dynamic distillation strategy is beneficial for our method in a black-box setting. 
Note that we only discuss the evaluation of the dynamic distillation for case B due to time constraints.

\begin{table}[t]
\caption{Average attack success rates in dynamic-distillation-based attacks}
 \begin{center}
 \begin{tabular}{l|cc} 
Domain & VGG(B) & ResNet(B) \\ \hline
hair color & 0.89 & 0.73 \\ \cline{1-3}
makeup 	& 0.82 & 0.69 \\ \cline{1-3}
eyeglasses & 0.84 & 0.71 \\ \hline 
 \end{tabular}
 \end{center}
\end{table}

\subsection{Human Perceptual Study}
We evaluated the quality of the images generated with our method using AMTurk to confirm that the images satisfy condition ($1$): 
they should be perceptually realistic enough to maintain the identifiability of the individual as an original image. 
We selected $100$ pairs of original and adversarial images and asked workers the question, “Do the two images have the same personal 
identity?” 
Note we assigned each pair to five different workers for fair comparison. 
For this question, $76.6\%$ of workers answered that personal identities in the original and adversarial images were the same. 
This result indicates that our method successfully translates images while maintaining the identifiability of the individual to avoid confusing human 
observers.

\subsection{A Comparison between Perturbations and Certified Radius}
To meet condition (2): our adversarial examples have large enough perturbations to bypass defenses based on small perturbations, 
we evaluated a state-of-the-art certified defense with randomized smoothing~\cite{cohen2019certified}. 
This defense provides certified regions where the classifier has constant output around each data point.

\textbf{Setup}. We constructed smoothed target face recognition model $\tilde{\phi}$ that return the most likely class when $x$ is 
perturbed as follows:
\begin{equation}
\tilde{\phi} = {\rm argmax}_{y \in \cal{Y}} P( \phi(x+\epsilon)=y),
\end{equation}
where $\epsilon$ is Gaussian noise with the standard deviation $\sigma$ and $\phi$ is a base classifier trained with Gaussian data 
augmentation~\cite{lecuyer2018certified}.

To compute the certified regions, we used Cohen's method with $n_{0}=100, n=1000,$ and $\alpha=0.001$~\cite{cohen2019certified}. 
Each certified region is represented by its radius.
The ball centered at each image with the radius is guaranteed to have constant output inside the ball.

We used VGG16 and ResNet50 trained with Gaussian data augmentation as the base classifiers for case A. 
The models were set to publically available pre-trained parameters as initial parameters and trained using stochastic 
gradient descent with $1000$ epochs. We set the learning rate and momentum to $0.0001$ and $0.9$, respectively.

\textbf{Result}. Figure~2 shows the certified accuracy of our smoothed target face recognition models with radius threshold $T$. 
The certified accuracy represents the proportion of images classified correctly and whose certified radius is less than $T$ for all 
images. In both VGG16 (left) and ResNet50 (right), all certified radii were less than 630. 
We calculated the average perturbations of our method in L2. 
The average perturbations for VGGFace(A) were 14,739, 461,290,430, and 3,589,136 for hair color, makeup, and eyeglasses, respectively. The average perturbations for ResNet(A) were 15,342, 461,290,700, and 3,589,382 for hair color, makeup, and eyeglasses, respectively. 
Note all pixels have values ranging from $0$ to $255$. 
We confirmed that all the images generated with our method provide larger perturbations than the certified radii.

These results indicate that the state-of-the-art certified defense might be insufficient for our method. 
This does not mean that the certified defense is broken since the threat model is different from ours.

\section{Conclusion}
We proposed a method for generating unrestricted adversarial examples against face recognition systems. 
The method translates the facial appearance of a source image into multiple domains to deceive target face recognition systems. 
Through experiments, we demonstrated that our method achieved about a $90\%$ attack success rate in a white-box setting with respect 
to several domains: hair color, makeups, and eyeglasses. 
We also evaluated our method in a black-box setting using transferability of adversarial examples and the dynamic distillation 
strategy, resulting in $30$ and $80\%$ attack success rates, respectively. 
We also demonstrated that perturbations introduced by our method were large enough to bypass a state-of-the-art certified defense, 
while the translation prevented the damages in the identifiability of the individual of a source image for human observers. 
We conclude that our method is promising for improving the robustness of face recognition systems.

\begin{figure*}[t]
\begin{center}
\includegraphics[width=1.2\columnwidth,natwidth=1500,natheight=1150]{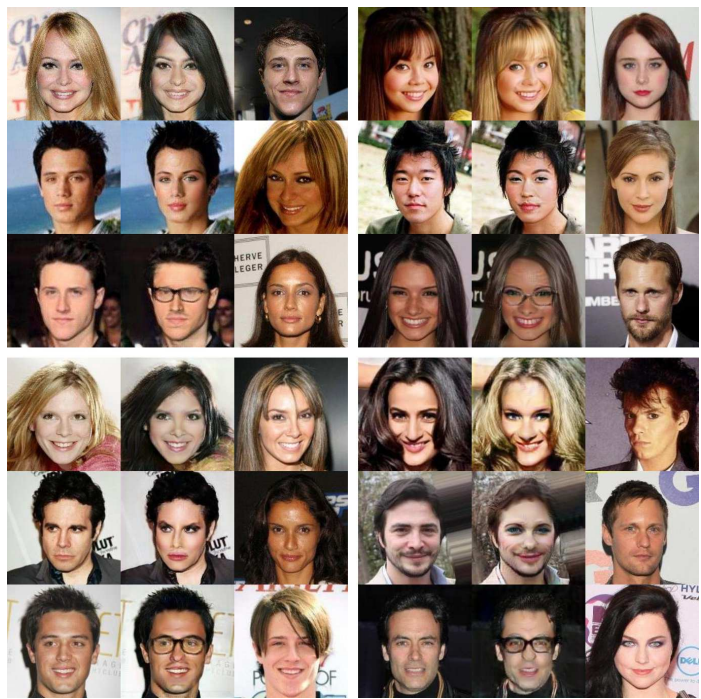}
\caption{Successful adversarial images. Left columns show original images and middle ones show successful adversarial examples. 
Corresponding target individuals are presented in right columns. 
Target domain labels are hair color (black or blond), makeup, and eyeglasses. 
Introduced perturbations were perceptible while preventing alteration of original identity.}
\label{fig:AX}
\end{center}
\end{figure*}

\bibliographystyle{aaai}
\bibliography{aaai2020} 

\begin{thebibliography}{}

\bibitem[\protect\citeauthoryear{Bahdanau, Cho, and
  Bengio}{2014}]{bahdanau2014neural}
Bahdanau, D.; Cho, K.; and Bengio, Y.
\newblock 2014.
\newblock Neural machine translation by jointly learning to align and
  translate.
\newblock {\em arXiv preprint arXiv:1409.0473}.

\bibitem[\protect\citeauthoryear{Brown \bgroup et al\mbox.\egroup
  }{2018}]{brown2018unrestricted}
Brown, T.~B.; Carlini, N.; Zhang, C.; Olsson, C.; Christiano, P.; and
  Goodfellow, I.
\newblock 2018.
\newblock Unrestricted adversarial examples.
\newblock {\em arXiv preprint arXiv:1809.08352}.

\bibitem[\protect\citeauthoryear{Cao \bgroup et al\mbox.\egroup }{2018}]{Cao18}
Cao, Q.; Shen, L.; Xie, W.; Parkhi, O.~M.; and Zisserman, A.
\newblock 2018.
\newblock Vggface2: A dataset for recognising faces across pose and age.
\newblock In {\em International Conference on Automatic Face and Gesture
  Recognition}.

\bibitem[\protect\citeauthoryear{Carlini and Wagner}{2017}]{carlini2017towards}
Carlini, N., and Wagner, D.
\newblock 2017.
\newblock Towards evaluating the robustness of neural networks.
\newblock In {\em Proceedings of the IEEE Symposium on Security and Privacy},
  39--57.

\bibitem[\protect\citeauthoryear{Choi \bgroup et al\mbox.\egroup
  }{2018}]{choi2018stargan}
Choi, Y.; Choi, M.; Kim, M.; Ha, J.-W.; Kim, S.; and Choo, J.
\newblock 2018.
\newblock Stargan: Unified generative adversarial networks for multi-domain
  image-to-image translation.
\newblock In {\em Proceedings of the IEEE Conference on Computer Vision and
  Pattern Recognition},  8789--8797.

\bibitem[\protect\citeauthoryear{Cisse \bgroup et al\mbox.\egroup
  }{2017}]{cisse2017parseval}
Cisse, M.; Bojanowski, P.; Grave, E.; Dauphin, Y.; and Usunier, N.
\newblock 2017.
\newblock Parseval networks: Improving robustness to adversarial examples.
\newblock In {\em Proceedings of the 34th International Conference on Machine
  Learning-Volume 70},  854--863.
\newblock JMLR. org.

\bibitem[\protect\citeauthoryear{Cohen, Rosenfeld, and
  Kolter}{2019}]{cohen2019certified}
Cohen, J.~M.; Rosenfeld, E.; and Kolter, J.~Z.
\newblock 2019.
\newblock Certified adversarial robustness via randomized smoothing.
\newblock In {\em Proceedings of the 36th International Conference on Machine
  Learning}.

\bibitem[\protect\citeauthoryear{Feng and
  Prabhakaran}{2013}]{feng2013facilitating}
Feng, R., and Prabhakaran, B.
\newblock 2013.
\newblock Facilitating fashion camouflage art.
\newblock In {\em Proceedings of the 21st ACM international conference on
  Multimedia},  793--802.

\bibitem[\protect\citeauthoryear{Goodfellow \bgroup et al\mbox.\egroup
  }{2014}]{goodfellow2014generative}
Goodfellow, I.; Pouget-Abadie, J.; Mirza, M.; Xu, B.; Warde-Farley, D.; Ozair,
  S.; Courville, A.; and Bengio, Y.
\newblock 2014.
\newblock Generative adversarial nets.
\newblock In {\em Advances in neural information processing systems},
  2672--2680.

\bibitem[\protect\citeauthoryear{Goodfellow, Shlens, and
  Szegedy}{2015}]{Ian2015expaining}
Goodfellow, I.; Shlens, J.; and Szegedy, C.
\newblock 2015.
\newblock Explaining and harnessing adversarial examples.
\newblock In {\em International Conference on Learning Representations}.

\bibitem[\protect\citeauthoryear{Goswami \bgroup et al\mbox.\egroup
  }{2018}]{goswami2018unravelling}
Goswami, G.; Ratha, N.; Agarwal, A.; Singh, R.; and Vatsa, M.
\newblock 2018.
\newblock Unravelling robustness of deep learning based face recognition
  against adversarial attacks.
\newblock In {\em Thirty-Second AAAI Conference on Artificial Intelligence}.

\bibitem[\protect\citeauthoryear{Gouk \bgroup et al\mbox.\egroup
  }{2018}]{gouk2018regularisation}
Gouk, H.; Frank, E.; Pfahringer, B.; and Cree, M.
\newblock 2018.
\newblock Regularisation of neural networks by enforcing lipschitz continuity.
\newblock {\em arXiv preprint arXiv:1804.04368}.

\bibitem[\protect\citeauthoryear{Gulrajani \bgroup et al\mbox.\egroup
  }{2017}]{gulrajani2017improved}
Gulrajani, I.; Ahmed, F.; Arjovsky, M.; Dumoulin, V.; and Courville, A.~C.
\newblock 2017.
\newblock Improved training of wasserstein gans.
\newblock In {\em Advances in Neural Information Processing Systems},
  5767--5777.

\bibitem[\protect\citeauthoryear{Isola \bgroup et al\mbox.\egroup
  }{2017}]{isola2017image}
Isola, P.; Zhu, J.-Y.; Zhou, T.; and Efros, A.~A.
\newblock 2017.
\newblock Image-to-image translation with conditional adversarial networks.
\newblock In {\em 2017 IEEE Conference on Computer Vision and Pattern
  Recognition},  5967--5976.

\bibitem[\protect\citeauthoryear{Karras \bgroup et al\mbox.\egroup
  }{2018}]{karras2018progressive}
Karras, T.; Aila, T.; Laine, S.; and Lehtinen, J.
\newblock 2018.
\newblock Progressive growing of gans for improved quality, stability, and
  variation.
\newblock In {\em International Conference on Learning Representations}.

\bibitem[\protect\citeauthoryear{Karras, Laine, and
  Aila}{2018}]{karras2018style}
Karras, T.; Laine, S.; and Aila, T.
\newblock 2018.
\newblock A style-based generator architecture for generative adversarial
  networks.
\newblock {\em arXiv preprint arXiv:1812.04948}.

\bibitem[\protect\citeauthoryear{Krizhevsky, Sutskever, and
  Hinton}{2012}]{krizhevsky2012imagenet}
Krizhevsky, A.; Sutskever, I.; and Hinton, G.~E.
\newblock 2012.
\newblock Imagenet classification with deep convolutional neural networks.
\newblock In {\em Advances in neural information processing systems},
  1097--1105.

\bibitem[\protect\citeauthoryear{Lecuyer \bgroup et al\mbox.\egroup
  }{2019}]{lecuyer2018certified}
Lecuyer, M.; Atlidakis, V.; Geambasu, R.; Hsu, D.; and Jana, S.
\newblock 2019.
\newblock Certified robustness to adversarial examples with differential
  privacy.
\newblock In {\em Proceedings of the IEEE Symposium on Security and Privacy},
  726--742.

\bibitem[\protect\citeauthoryear{Li \bgroup et al\mbox.\egroup
  }{2018}]{li2018second}
Li, B.; Chen, C.; Wang, W.; and Carin, L.
\newblock 2018.
\newblock Second-order adversarial attack and certifiable robustness.
\newblock {\em arXiv preprint arXiv:1809.03113}.

\bibitem[\protect\citeauthoryear{Liu \bgroup et al\mbox.\egroup
  }{2015}]{liu2015faceattributes}
Liu, Z.; Luo, P.; Wang, X.; and Tang, X.
\newblock 2015.
\newblock Deep learning face attributes in the wild.
\newblock In {\em Proceedings of International Conference on Computer Vision}.

\bibitem[\protect\citeauthoryear{Masi \bgroup et al\mbox.\egroup
  }{2018}]{masi2018deep}
Masi, I.; Wu, Y.; Hassner, T.; and Natarajan, P.
\newblock 2018.
\newblock Deep face recognition: a survey.
\newblock In {\em 2018 31st SIBGRAPI Conference on Graphics, Patterns and
  Images},  471--478.

\bibitem[\protect\citeauthoryear{Odena, Olah, and
  Shlens}{2017}]{odena2017conditional}
Odena, A.; Olah, C.; and Shlens, J.
\newblock 2017.
\newblock Conditional image synthesis with auxiliary classifier gans.
\newblock In {\em Proceedings of the 34th International Conference on Machine
  Learning},  2642--2651.

\bibitem[\protect\citeauthoryear{Papernot \bgroup et al\mbox.\egroup
  }{2016}]{papernot2016practical}
Papernot, N.; McDaniel, P.; Goodfellow, I.; Jha, S.; Celik, Z.~B.; and Swami,
  A.
\newblock 2016.
\newblock Practical black-box attacks against deep learning systems using
  adversarial examples.
\newblock {\em arXiv preprint arXiv:1602.02697} 1(2):3.

\bibitem[\protect\citeauthoryear{Parkhi, Vedaldi, and
  Zisserman}{2015}]{Parkhi15}
Parkhi, O.~M.; Vedaldi, A.; and Zisserman, A.
\newblock 2015.
\newblock Deep face recognition.
\newblock In {\em British Machine Vision Conference}.

\bibitem[\protect\citeauthoryear{Sharif \bgroup et al\mbox.\egroup
  }{2016}]{sharif2016accessorize}
Sharif, M.; Bhagavatula, S.; Bauer, L.; and Reiter, M.~K.
\newblock 2016.
\newblock Accessorize to a crime: Real and stealthy attacks on state-of-the-art
  face recognition.
\newblock In {\em Proceedings of the 2016 ACM SIGSAC Conference on Computer and
  Communications Security},  1528--1540.

\bibitem[\protect\citeauthoryear{Sharif \bgroup et al\mbox.\egroup
  }{2017}]{sharif2017adversarial}
Sharif, M.; Bhagavatula, S.; Bauer, L.; and Reiter, M.~K.
\newblock 2017.
\newblock Adversarial generative nets: Neural network attacks on
  state-of-the-art face recognition.
\newblock {\em arXiv preprint arXiv:1801.00349}.

\bibitem[\protect\citeauthoryear{Song \bgroup et al\mbox.\egroup
  }{2018}]{NIPS2018_8052}
Song, Y.; Shu, R.; Kushman, N.; and Ermon, S.
\newblock 2018.
\newblock Constructing unrestricted adversarial examples with generative
  models.
\newblock In {\em Advances in Neural Information Processing Systems}.
\newblock  8322--8333.

\bibitem[\protect\citeauthoryear{Szegedy \bgroup et al\mbox.\egroup
  }{2014}]{Szegedy2014IntriguingPO}
Szegedy, C.; Zaremba, W.; Sutskever, I.; Bruna, J.; Erhan, D.; Goodfellow,
  I.~J.; and Fergus, R.
\newblock 2014.
\newblock Intriguing properties of neural networks.
\newblock In {\em International Conference on Learning Representations}.

\bibitem[\protect\citeauthoryear{Tsuzuku, Sato, and
  Sugiyama}{2018}]{NIPS2018_7889}
Tsuzuku, Y.; Sato, I.; and Sugiyama, M.
\newblock 2018.
\newblock Lipschitz-margin training: Scalable certification of perturbation
  invariance for deep neural networks.
\newblock In {\em Advances in Neural Information Processing Systems}.
\newblock  6542--6551.

\bibitem[\protect\citeauthoryear{Xiao \bgroup et al\mbox.\egroup
  }{2018}]{ijcai2018-543}
Xiao, C.; Li, B.; yan Zhu, J.; He, W.; Liu, M.; and Song, D.
\newblock 2018.
\newblock Generating adversarial examples with adversarial networks.
\newblock In {\em Proceedings of the Twenty-Seventh International Joint
  Conference on Artificial Intelligence, {IJCAI-18}},  3905--3911.

\bibitem[\protect\citeauthoryear{Zhu \bgroup et al\mbox.\egroup
  }{2017}]{zhu2017unpaired}
Zhu, J.-Y.; Park, T.; Isola, P.; and Efros, A.~A.
\newblock 2017.
\newblock Unpaired image-to-image translation using cycle-consistent
  adversarial networks.
\newblock In {\em IEEE International Conference on Computer Vision},
  2242--2251.

\end{thebibliography}
  
\end{document}